\documentclass[10pt,twocolumn,letterpaper]{article}

\usepackage{iccv}     

\usepackage{array}
\usepackage{colortbl}

\usepackage{multirow}
\usepackage{bbding}  
\usepackage{makecell} 
\usepackage{amssymb} 
\usepackage{bm}
\usepackage{textcomp} 
\usepackage{enumitem} 
\usepackage{threeparttable}   
	\usepackage{booktabs}%

\definecolor{iccvblue}{rgb}{0.21,0.49,0.74}
\usepackage[pagebackref,breaklinks,colorlinks,allcolors=iccvblue]{hyperref}

\def\paperID{11326} 
\def\confName{ICCV}
\def\confYear{2025}

\title{Event-based Tiny Object Detection: A Benchmark Dataset and Baseline}

\author{Nuo Chen, Chao Xiao, Yimian Dai, Shiman He, Miao Li, Wei An}

\begin{document}
\maketitle
\begin{abstract}
Small object detection (SOD) in anti-UAV  task is a challenging problem due to the small size of UAVs and complex backgrounds. Traditional frame-based cameras struggle to detect small objects in complex environments due to their low frame rates, limited dynamic range, and data redundancy. Event cameras, with microsecond temporal resolution and high dynamic range, provide a more effective solution for SOD. However, existing event-based object detection datasets  are limited in scale, feature large targets size, and lack diverse backgrounds, making them unsuitable for SOD benchmarks. In this paper, we introduce a Event-based Small object detection (EVSOD) dataset (namely EV-UAV), the first large-scale, highly diverse benchmark for anti-UAV tasks. It includes 147 sequences with over 2.3 million event-level annotations, featuring extremely small targets (averaging 6.8 $\times$ 5.4 pixels) and diverse scenarios such as urban clutter and extreme lighting conditions. Furthermore, based on the observation that small moving targets form continuous curves in spatiotemporal event point clouds, we propose Event based Sparse Segmentation Network (EV-SpSegNet), a novel baseline for event segmentation in point cloud space, along with a Spatiotemporal Correlation (STC) loss that leverages motion continuity to guide the network in retaining target events. Extensive experiments on the EV-UAV dataset demonstrate the superiority of our method and provide a benchmark for future research in EVSOD. The dataset and code are at \url{https://github.com/ChenYichen9527/Ev-UAV}.
 
\end{abstract}    

\section{Introduction}
\label{sec:intro}
Unmanned Aerial Vehicles (UAVs) have made significant contributions to fields such as geological exploration~\cite{uav} and precision agriculture~\cite{uav2} due to their high mobility and low cost. However, the rapid increase in UAV numbers has also introduced potential threats to public safety, making anti-UAV technologies increasingly critical. Small Object Detection (SOD), which aims to accurately and real-time locate tiny targets, is fundamental to anti-UAV tasks. However, it remains a challenging problem due to the small size of UAVs and the complexity of their surrounding environments.
\begin{figure}[!t]
	\centering
	\includegraphics{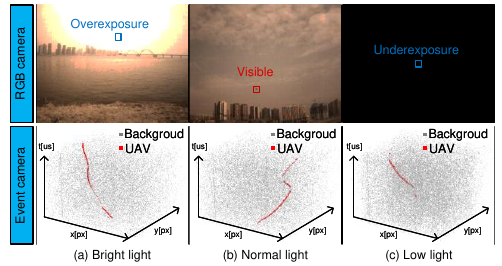}
	\caption{Comparison between event camera and RGB camera.  The RGB camera can only capture the objects under normal light, while the event camera  can capture objects under various extreme lighting conditions. And the event camera can capture the continuous motion  trajectory of the small object (shown as the red curve).} 
	\label{top_right}
\end{figure}

Conventional frame-based cameras suffer from limited frame rates (typically 30-60 Hz), narrow dynamic range, and redundant data capture, hindering the development of SOD~\cite{huang2023anti,jiang2021anti}.
Event cameras~\cite{posch2010qvga,gallego2020event}, bio-inspired sensors that asynchronously capture per-pixel intensity changes with microsecond temporal resolution ($ \ge {10^6}$ Hz) and high dynamic range (\(120\)dB), offer a promising alternative to SOD under real-world scenarios. Their ability to encode motion as sparse spatio-temporal event streams inherently avoids data redundancy and preserves fine-grained details of high-speed objects, even in challenging illumination.

Despite this potential, the following challenges  hinder their adoption in SOD: 
First, the small size and limited appearance cues of small objects hinder feature learning, while background clutter degrades detection performance and increases false alarms. 
Second, current event-based methods~\cite{rvt,1Mpx} often convert events into synchronized image representations for frame-based detection. This disrupts the continuity  and sparsity  of event data, causing loss of temporal information and unnecessary processing of redundant background data.
Furthermore, the lack of large-scale, high-quality datasets significantly impedes the development of event-based small moving object detection.  

Therefore, in this paper, we aim to address the aforementioned challenges to advance the development of the event-based small moving object (especially for UAVs) detection (EVSOD) by firstly building a large-scale dataset for EVSOD, then developing a novel detection method tailored for EVSOD based on the spatio-temproal event point cloud representation, and finally extensively evaluating various deep feature learning methods for EVSOD with the developed dataset and the proposed method as baseline.

Specifically, we construct a new dataset for EVSOD, for the first time, to address the following issues:  (1) Existing SOD datasets primarily focus on  visible modality~\cite{vis1,vis2,vis3}, infrared modality~\cite{inf1,inf2}, or  the fusion of these two modalities~\cite{yxy}, and few research is conducted to explore the EVSOD. (2) Although several event-based object detection datasets~\cite{rvt,1Mpx,eventvot} have been proposed, they focus on detecting large objects in simple backgrounds, failing to capture real-world scenarios where small objects appear in cluttered environments. (3) Existing event-based object detection datasets~\cite{f-uav-det,nerrd} lack event-level annotations, restricting the development of algorithms using other event representations (e.g., sparse point clouds and SNNs). The aforementioned issues urge us to introduce the first large-scale dataset with event-level annotations for EVSOD, advancing the development of anti-UAV.

Then, we propose Event based Sparse Segmentation Network (EV-SpSegNet), which specially designed for efficient and robust EVSOD.  We observe that small moving objects in spatiotemporal event point clouds form continuous, elongated curves, which are distinct from the surfaces formed by the background and the discrete points formed by noise. Leveraging this characteristic, we design a Grouped Dilated Sparse Convolution Attention (GDSCA) module to extract multi-scale local and long-term contextual features of curve structures. Based on the proposed GDSCA, we construct our network to detect small moving objects. Furthermore, we design a new Spatiotemporal Correlation (STC) loss, which evaluates the local spatiotemporal correlation of events to guide the network in retaining more events with continuous curves.

Finally, we conduct extensive performance evaluations on the proposed dataset, comparing our method with 13 state-of-the-art algorithms, including 4 frame-based methods~\cite{ssd,fasterrcnn,detr,yolov10}, 6 event-based methods~\cite{rvt,sast,get,1Mpx,emsyolo,spikeyolo}, and 3 point cloud segmentation methods~\cite{kpconv,randla,coseg}, enabling fair and reproducible comparisons for EVSOD.

The main contributions are summarized as follows:
(1) We  build the first large-scale dataset \textbf{EV-UAV}, with high diversity for EVSOD. It contains 147 event sequences with event-level annotations, covering challenging scenarios like high-brightness and low-light conditions, with targets averaging 1/50 the size in existing datasets.
(2) We propose\textbf{ EV-SpSegNet}, a novel detection framework that distinguishes small targets from background and noise based on the continuous curves formed by targets. Additionally, we  introduce \textbf{STC loss}, which retains target  events with continuous motion through local spatiotemporal correlation.
(3) Based on the proposed dataset, we conduct a comprehensive evaluation of our method and 13 state-of-the-art approaches, demonstrating the superiority of our method and providing a benchmark for future research in EVSOD.

\section{Related work}
\label{sec:Related work}

\begin{table*}[ht]
	\caption{Summary of existing event UAV detection datasets. BL: bright light, NL: normal light, LL: low light, MS: multi-scene, MT: multi-target.}
	\footnotesize
	\label{tab:uav_dataset}
	\centering
	\begin{tabular}{c|c|c|c|p{0.7cm}<{\centering}|ccc|cc|c}
		\hline
		\multirow{2}{*}{\textbf{Dataset}} & \multirow{2}{*}{\textbf{\#AGV.UAV scale}} & \multirow{2}{*}{\textbf{Label Type}} & \multirow{2}{*}{\textbf{\makecell{ UAV Sequence\\Ratio($\%$) }}} & \multirow{2}{*}{\textbf{\makecell{UAV \\ centric}}} & \multicolumn{3}{c|}{\textbf{Lighting conditions}} & \multicolumn{2}{c|}{\textbf{Object}} & \multirow{2}{*}{\textbf{Year}} \\ \cline{6-10} 
		&                                   &                               &                             &                              & \textbf{BL} & \textbf{NL} & \textbf{LL} & \textbf{MS} & \textbf{MT} & \\ \hline
		VisEvent~\cite{wang2023visevent}                 &      84×66 pixels                              & BBox                          &  15.97                          & $\bm{\times}$                  & $\bm{\times}$     & \textbf{\textcolor{red}{\checkmark}} & \textbf{\textcolor{red}{\checkmark}}    & \textbf{\textcolor{red}{\checkmark}} & $\bm{\times}$ & 2023 \\
		EventVOT~\cite{eventvot}                     &     129×100 pixels                             & BBox                          &  8.41                           & $\bm{\times}$                  & $\bm{\times}$     & \textbf{\textcolor{red}{\checkmark}} & \textbf{\textcolor{red}{\checkmark}}    & \textbf{\textcolor{red}{\checkmark}} & $\bm{\times}$ & 2024 \\
		EvDET200K~\cite{wang2024object}                &  68×45 pixels                                 & BBox                          &   3.57                          &$\bm{\times}$    & $\bm{\times}$     & \textbf{\textcolor{red}{\checkmark}} & \textbf{\textcolor{red}{\checkmark}}    & \textbf{\textcolor{red}{\checkmark}} & \textbf{\textcolor{red}{\checkmark}}  & 2024 \\
		F-UAV-D~\cite{f-uav-det}                  & -                                  & BBox                          & 100                     & \textbf{\textcolor{red}{\checkmark}}                   & $\bm{\times}$     & \textbf{\textcolor{red}{\checkmark}} & $\bm{\times}$    & $\bm{\times}$ & $\bm{\times}$ & 2024 \\
		NeRDD~\cite{nerrd}                    &  55×31 pixels                                & BBox                          &  100                          & \textbf{\textcolor{red}{\checkmark}}    & $\bm{\times}$     & \textbf{\textcolor{red}{\checkmark}} & $\bm{\times}$    & $\bm{\times}$ & \textbf{\textcolor{red}{\checkmark}}  & 2024 \\
		\hline 
		\rowcolor{green!15} EV-UAV & \textbf{6.8×5.4 pixels}                   & \textbf{Seg}                           & \textbf{100}                        & \textbf{\textcolor{red}{\checkmark}}    & \textbf{\textcolor{red}{\checkmark}} & \textbf{\textcolor{red}{\checkmark}} & \textbf{\textcolor{red}{\checkmark}} & \textbf{\textcolor{red}{\checkmark}} & \textbf{\textcolor{red}{\checkmark}} & 2025 \\ \hline
	\end{tabular}
\end{table*}
\subsection{Anti-UAV detection dataset}
Currently, event-based object detection datasets  primarily focus on autonomous driving~\cite{gen1,1Mpx} and general object detection~\cite{ev_obj_det0,ev_obj_det1,ev_obj_det2}.  There is limited attention given to  datasets that are  exclusively designed for UAV detection. In~\cref{tab:uav_dataset}, we provide a comprehensive summary of existing datasets, highlighting the scarcity of event-based datasets for UAV object detection.

Early datasets often included UAVs only as a minor subcategory within larger datasets~\cite{ev_obj_det0,ev_obj_det1,ev_obj_det2}.  Due to their limited scale and diversity, these datasets are insufficient for training modern deep learning-based methods.  The recently proposed F-UAV-D dataset~\cite{f-uav-det} is specifically designed for event-based UAV object detection. However, it only covers indoor environments, which limits its applicability.
NeRDD~\cite{nerrd}, the first dataset dedicated to event-based UAV  detection, considers only large objects against clean backgrounds. This fails to capture real-world scenarios where UAVs often appear as small objects in dynamic and cluttered backgrounds.
To address these limitations, we introduce EV-UAV, a dataset designed to better represent challenging  real-world scenarios. EV-UAV features a large number of detection scenes with small targets and dynamic backgrounds, providing a robust benchmark for developing and evaluating anti-UAV methods.

\subsection{Object detection for event cameras}
Existing object detection methods for event cameras can be broadly divided into two  categories. The first category focuses on directly processing sparse event data. For example, graph-based methods model events as dynamic spatiotemporal graphs~\cite{gnn0,gnn1,gnn2}, restricting information propagation to a finite set of sparse connections to reduce computational complexity. Similarly, Spiking Neural Network (SNN)-based methods process asynchronous and sparse event streams directly within the network architecture~\cite{snn1,snn2,snn3,snn4}. By transmitting only discrete spike signals between neurons, these methods significantly reduce power consumption during training. However, sparse representation-based approaches are limited by their reliance on specialized hardware or suboptimal performance in complex scenarios.

The second category transforms sparse event streams into image-like representations, enabling compatibility with dense convolutional operations. Early dense-based methods  directly used event representations generated from short time windows for detection~\cite{3dtensorl,eventcamera2,benosman2013event,UnsupervisedEVFlownet}. These methods, however,  only considered event information within a limited time, resulting in limited performance. Later work addressed this issue by incorporating recurrent neural networks~\cite{astmnet,1Mpx,rvt}, which improved detection performance by capturing temporal dependencies. Despite these advancements, such methods consume significant computational resources for event transformation and processing artificially introduced redundant background data, while struggling to detect small targets due to their lack of distinct texture and contour features.
In this work, we observe that small targets form continuous curves in spatiotemporal event point clouds, which are distinct from background and noise. Leveraging this characteristic, we design EV-SpSegNet, which directly segments continuous target trajectories from sparse point clouds. This approach avoids the computational overhead caused by converting events into image-like representations. Moreover, the distinct curve features help the network better detect small targets.

\section{EV-UAV benchmark}

\subsection{Preliminaries}
Event cameras work totally  differently from frame-based cameras with fixed frame rates.  Specifically, each pixel of an event camera independently responds to changes in light intensity.  When the logarithmic brightness change reaches a predefined threshold, an event  $E=(x,y,t,p)$ is triggered.  In this tuple, $x$ and $y$ represent the pixel coordinates, $p$  indicates the polarity of the brightness change (positive or negative), and $t$ is the timestamp of the event.

Unlike previous annotations for object detection~\cite{rvt,1Mpx} or semantic segmentation~\cite{ess,evsegnet,dsec}, we do not convert events into synchronized frames.  Instead, we segment events directly in the point cloud space, event by event.
The sources of events can be categorized into three types: events generated by the target (${E_T}$), events generated by the background (${E_B}$), and events caused by noise (${E_N}$). This can be represented as:  $E = {E_T} + {E_B} + {E_N}$. Our goal is to segment all events generated by target motion from the raw event stream. The final annotation is a set of discrete events ${E_{label}} \in {E_T}$.

\subsection{Data Collection and Annotations}

\subsubsection{Data Collection}
\begin{figure*}[t]
	\centering
	\includegraphics{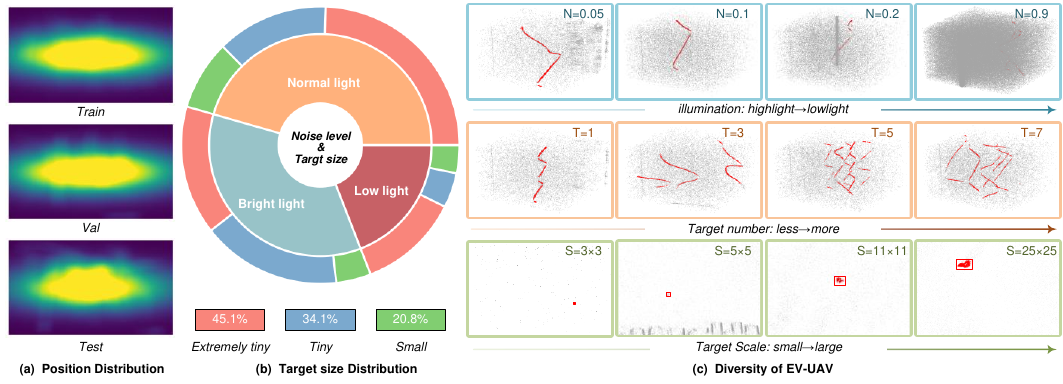}\vspace{-0.1cm}
	\caption{Illustration of the proposed dataset EV-UAV. (a) Position distribution of EV-UAV. (b) Distribution of noise level and illumination of the proposed dataset. (c) EV-UAV collects diverse event streams across various conditions of illumination and target number and scale. N represents the noise level, reflecting the proportion of events in spatiotemporal space. T denotes the number of targets. S indicates the size of targets in the image plane.}\vspace{-0.1cm}
	\label{dataset}
\end{figure*}

\begin{figure}[t]
	\centering
	\includegraphics{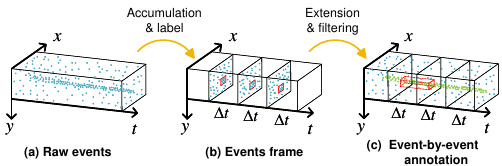}\vspace{-0.1cm}
	\caption{Event-by-event annotation method. Firstly, the events are accumulated into event frames, and targets are annotated with 2D bounding boxes on the event frames. Then, according to the time interval $\Delta t$ corresponding to each frame, the 2D annotations on the image plane are extended into 3D bounding box annotations in XYT 3D space, and  all events within the 3D boxes are filtered  as targets.}\vspace{-0.4cm}
	\label{label}
\end{figure}

We collect data using a DAVIS346 camera mounted on a tripod to monitor a fixed range of scenes.  This camera has a spatial resolution of 346$\times$260 and a temporal resolution of up to $10^6$ HZ. Our data collection spans from May 1, 2023, to June 1, 2024, covering various times of the day, seasons, and weather conditions.  To accurately reflect the movement patterns of UAVs and the complexity of real-world scenarios, we capture UAVs performing diverse actions in various environments. These actions include forward and backward motion, lateral movements, ascending and descending maneuvers, as well as complex combinations of these behaviors.  The dataset encompasses  three different lighting conditions (normal, bright  and low light) and various backgrounds, such as buildings, mountains, forests, urban areas, clouds, water surfaces, etc. We carefully annotated 147 sequences of data, totaling over 20.3 million target  events.

\subsubsection{Event by Event Annotation}
Due to the asynchronous and sparse output of event cameras, manually annotating per-event data  is highly time-consuming and inefficient. Existing event camera datasets are typically  recorded using a combination of event cameras and standard RGB cameras. Labels are first extracted from the frame-based cameras and then transferred to the pixel coordinates of the event cameras through  geometric transformations. 
However, this labeling strategy is limited to producing low frame rate bounding box annotations, which cannot match the high-frequency output characteristics of event cameras or meet the fine-grained, event-by-event labeling requirements.
Although manual event-by-event labeling could theoretically improve annotation accuracy, the process is highly labor-intensive and time-consuming, making it impractical for real-world applications. To address this issue, we propose a novel method that generates approximate event-by-event annotations using bounding box labels, as illustrated in the Fig.~\ref{label}. This approach not only achieves fine-grained event-level annotations, enabling microsecond-level object detection, but also allows for degradation into bounding box annotations at any given time, thus supporting the learning requirements of some asynchronous event-based object detection algorithms.

\subsection{Benchmark Features and Statistics}
\begin{figure*}[t]
	\centering
	\includegraphics{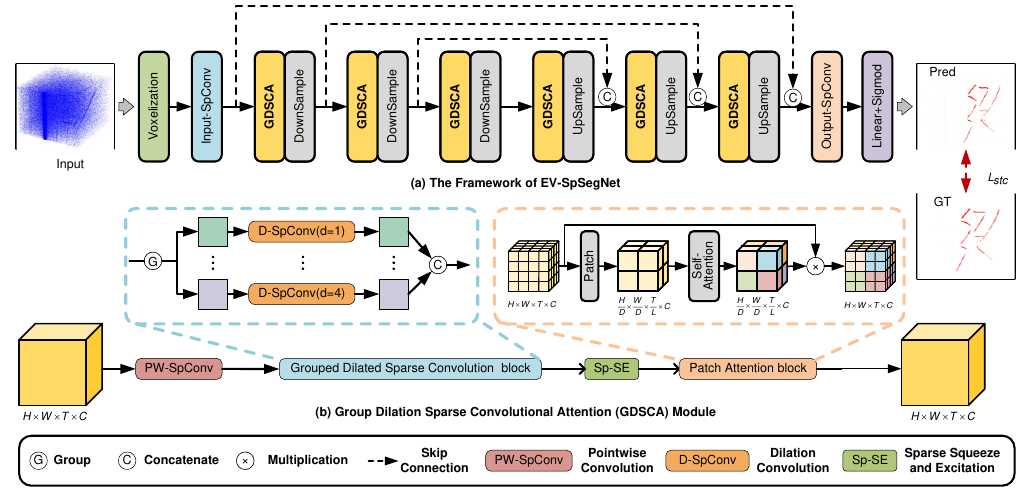}
	\caption{ (a) The overview of the proposed Ev-SpSegNet. The network  is built on a U-shaped architecture consisting of symmetric encoder and decoder.  (b) The GDSCA module. The Grouped Dilated Sparse Convolution block is applied to features from different channels to extract local temporal features at multiple scales. The Sp-SE module is used to integrate these features. And the Patch Attention block is used to downsample voxel features and enable global context feature interaction.}
	\label{fig:netwok}
\end{figure*}
\textbf{Dataset Splitting.} The EV-UAV dataset is divided into training (99 sequences), validation (24 sequences), and test sets (24 sequences), totaling 147 event sequences.

\textbf{Position Distribution.} As shown in \cref{dataset}(a), the targets are mostly located in the central area of the image. The larger horizontal variance suggests that the drone's movement is predominantly horizontal. Additionally, the range of target movements in the validation set is more varied. The fluctuations in the validation set are greater than those in the test set in corresponding directions.

\textbf{Small-Scale Targets.} In real-world scenarios, due to the long imaging distance and the small size of UAVs, drone targets appear at very small scales in the images. The general scale ranks\footnote{ small$\in$[1, ${32^2}$), medium$\in$[${32^2}$, ${96^2}$), and large$\in$[${96^2}$, $\infty$).} of \cite{sizerank} are no longer applicable.  Following previous work~\cite{yxy}, we further divide small targets into three levels: extremely tiny$\in$[1, ${8^2}$), tiny$\in$[${8^2}$, ${16^2}$), and small$\in$[${16^2}$, $\infty$). Fig.~\ref{dataset}(b) shows the number of annotations for each target category. It can be observed that   the scale of all targets is smaller than ${32^2}$, with extremely tiny targets accounting for the largest proportion, approximately 45$\%$.
 Compared to other UAV detection datasets, our EV-UAV dataset has the smallest average UAV scale, as indicated in \cref{tab:uav_dataset}.

\textbf{Rich Diversity.} To represent real-world anti-UAV target detection scenarios, we endow EV-UAV with diversity. This helps the dataset cover various complex situations in anti-UAV detection both in time and space. As shown in Fig.~\ref{dataset}(c) , the diversity of EV-UAV is reflected not just in illumination levels but also in background complexity, target quantity, and scale.

\subsection{Evaluation Metrics}
In this work, we evaluate the event-by-event segmentation performance of the algorithm using Intersection over Union ($IoU$) and accuracy ($ACC$), and assess the  localization  performance using  probability of detection ($P_{d}$) and false alarm rate ($F_{a}$).
More details regarding evaluation metrics can be found in the supplementary material.

\section{Method}
Frame-based detectors require converting events into synchronized image-like representations, which compresses the temporal information of events and makes it even more difficult to detect small targets that already lack distinct appearance cues. In raw event data, events generated by moving targets exhibit strong spatiotemporal correlations, forming elongated continuous curves in event point cloud space. Leveraging this characteristic, we propose EV-SpSegNet, which directly segments events generated by moving targets in sparse event point clouds. Additionally, we introduce a novel spatiotemporal correlation loss that evaluates the local spatiotemporal correlation of events to guide the network in retaining target events with continuous curves.

\subsection{EV-SpSegNet}
Small moving targets form distinct continuous curves in spatiotemporal event point clouds. However, networks struggle to extract their features because these curves span long temporal ranges, and traditional convolution operations with limited receptive fields cannot effectively capture global features. While attention mechanisms can extract global context, their quadratic computational complexity restricts their ability to process large-scale event data. To effectively extract features from continuous curves in spatiotemporal event point clouds, we design a Grouped Dilated Sparse Convolution Attention (GDSCA) module, as shown in Fig.~\ref{fig:netwok}(b). This module first captures local multi-scale features through the Grouped Dilated Sparse Convolution (GDSC) Block and then enables global feature interaction across patches using patch attention. Specifically, the GDSC Block groups input features along the channel dimension, with each group applying convolutions of different dilation rates to extract multi-scale local features. These features are then fused in the Sparse Squeeze and Excitation (Sp-SE) block. Finally, the patch attention block divides the point cloud into larger sub-regions and implements global feature interaction between sub-regions using self-attention. This allows the network to model the differences between targets and background noise in greater detail, enhancing its ability to distinguish moving targets.

\subsection{Spatiotemporal Correlation Loss}
In general image segmentation tasks, the loss function used for training the network is typically the standard pixel-wise binary cross-entropy (BCE) loss:
\begin{equation}
	{L_{bce}}(p,y) = \left\{ {\begin{array}{*{20}{c}}
			{ - \log (p),y = 1}\\
			{ - \log (1 - p),y = 0}
		\end{array}.} \right.
	\label{eq:bceloss}
\end{equation}

\begin{figure}[t]
	\centering
	\includegraphics{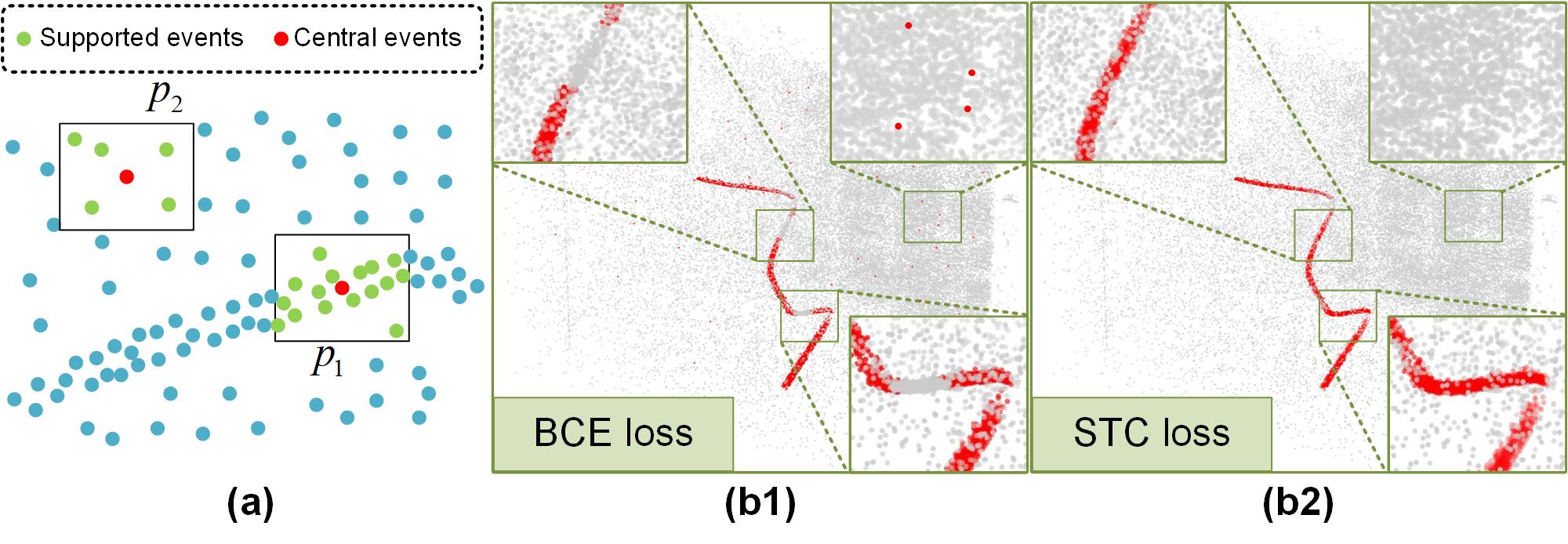}\vspace{-0.1cm}
	\caption{(a) The principle of spatiotemporal correlation loss. (b) Small object segmentation results of BCE loss and STC loss.}
	\label{fig:loss}
\end{figure}

Although EV-SpSegNet computes structured outputs and focuses on large neighborhood information to better utilize long-term features, losses like BCE loss process each pixel independently.
They do not consider the structure of neighborhoods, such as the spatiotemporal continuity of objects, which is crucial for describing the continuous curves formed by objects. 
As shown in Fig.~\ref{fig:loss}(a), compared to the isolated noise   ${p_2}$,   the event  ${p_1}$ with more high-confidence supporting events is more likely to be a true target. However, misclassifying both of them only results in a low coat of BCE loss.

To address this issue, we introduce a spatiotemporal correlation loss that encourages the network to retain more events with high spatiotemporal correlation while discarding more isolated noise.
Specifically, this loss  calculates the sum of the confidence of all supporting events within a $k \times k \times\tau$ neighborhood around the events, and defines it as the spatiotemporal correlation weight. We incorporate this weight into the cross-entropy loss and the spatiotemporal correlation loss is defined as follows:
\begin{equation}
{L_{stc}}(p,y) = \left\{ {\begin{array}{*{20}{c}}
		{ - {{w_{stc}}^\gamma }\log (p),y = 1}\\
		{ - {(1 - {w_{stc}})}^{^\gamma }\log (1 - p),y = 0}
\end{array}}, \right.
	\label{eq:Lstc}
\end{equation}
where the  $y \in \{ 0,1\}$ and ${p} \in (0,1)$ represent the segmentation label and prediction result of the event, respectively. $\gamma$ is a hyperparameters to control the influence of spatiotemporal correlations. ${w_{stc}}$ is the spatiotemporal correlation weight of each event, which is defined as: 
\begin{equation}
	{w_{stc}}(p) = sigmoid(\sum\limits_{{e^j} \in {V^{k\tau }}} {{p^j}} ),
	\label{eq:wstc}
\end{equation}
where  $V^{k\tau }$ represents the neighborhood of events with a size of  $k \times k \times \tau $.
As illustrated in Fig.~\ref{fig:loss} (b1-b2), the STC loss preserves the continuous event trajectories generated by moving objects and eliminates false alarms caused by misclassification.

\section{Experiments}
\subsection{Experimental Settings}
We used  8-second events as input and set the size of each voxel to 1 pixel $\times$ 1 pixel  $\times$ 1 ms. The batch size was set to 1, and all layers of the network were randomly initialized. We trained our network using the Adam optimizer~\cite{adam}   for 50 epochs  with the learning rate linearly decaying from 1e-2 to 1e-3.
The model was trained  on the predefined training set of EV-UAV dataset, and we chose the models based on their best performance on the validation set and evaluate them on the test set. The hyperparameters $k$, $\tau$, and $\gamma$ of the STC loss are set to 3, 5, and 2, respectively. All models were implemented with PyTorch on a single NVIDIA RTX 3090 GPU. 

\begin{table*}[t]
	\begin{threeparttable}
		\footnotesize
		\caption{Quantitative comparison of the proposed method to state-of-the-art methods. }
		\centering
		\begin{tabular}{lccccccccc}
			\hline
			Methods     & Publications & Event Rep.   & Temporal & $IoU$(\%)↑ & $ACC$(\%)↑ & ${P_d}$(\%)↑ & ${F_a}$(10\textasciicircum{}-4)↓ & \#Params. & Runtime(ms) \\ \cline{1-10}
			SSD~\cite{ssd}         & ECCV 2016             & Event Count  & No       & 25.31    & 28.56    & 26.31   & 486.63                      & 25.2M    & 2113             \\
			Faster RCNN~\cite{fasterrcnn} & TPAMI 2016             & Event Count  & No       & 26.93    & 29.68    & 27.39   &  689.68                      & 41.2M     & 3962             \\
			DETR~\cite{detr}        & ICLR 2020    & Event Count  & No      & 30.35    & 33.63    & 31.64   & 631.37                      & 39.8M      & 3136            \\
			YOLOV10-S~\cite{yolov10}       &  NIPS 2025   & Event Count  & No      & 32.55    & 33.39    & 32.18   & 589.67                      & 7.3M     & 1627            \\
			
			\hline
			EMS-YOLO~\cite{emsyolo}    & ICCV 2023    & SNN          & Yes       & 36.77    & 42.92    & 50.68   & 112.36                 & \underline{3.3M}    & 1229            \\
			Spike-YOLO~\cite{spikeyolo}  & ECCV 2024    & SNN          & Yes     & 43.94    & 48.26    & 59.62   & 55.38                              & 69.0M     & 1883            \\
				GET~\cite{get}         & ICCV 2023    & Groupv Token & Yes     & 40.31    & 48.91    & 60.73   & 46.35           & 18.4M     & 2168            \\
		 RED~\cite{1Mpx}         & NeurIPS 2020     & Voxel Grid   & Yes      & 35.99    & 45.54    & 53.76   & 102.27                       & 24.1M     & 3427             \\

				RVT~\cite{rvt}      &CVPR 2023    & Voxel Grid        & Yes     & 43.21    & 51.38    & 60.35   & 55.68                                    & 9.9M      & 1737            \\
			SAST~\cite{sast}        & CVPR 2024    & Voxel Grid   & Yes      & 34.31    & 40.22    & 51.21   & 150.32                 & 18.5M      &  3075           \\ \hline
			KPConv~\cite{kpconv}      & ICCV 2019    & Points       & Yes      & 48.19    & 57.28    & 68.59   & 16.32                       & 50.1M     & 562            \\
			RandLA-Net~\cite{randla}  & CVPR 2020    & Points       & Yes      & 50.32    & 59.29    & 70.56   & \underline{6.95}                        & \textbf{1.2M}     & 353           \\
			COSeg~\cite{coseg}         & CVPR 2024             &Points              & Yes         & \underline{51.89}         &\underline{60.93}          &\underline{71.32}         &9.21                            & 23.4M          & 364            \\
			\rowcolor{green!15} \textbf{Ours}        & -            & Points       & Yes      & \textbf{55.18}    & \textbf{65.02}    &\textbf{ 77.53}   & \textbf{1.63}                        & 4.0M      & \textbf{35.9}        \\ \hline
		\end{tabular}
		\label{tab:sota}
		\begin{tablenotes}    
			\footnotesize               
			\item  The \textbf{bold} and the \underline{underline}  represent the best and second-best performance, respectively.
		\end{tablenotes}  
	\end{threeparttable}
\end{table*}

\begin{figure*}[t]
	\centering
	\includegraphics{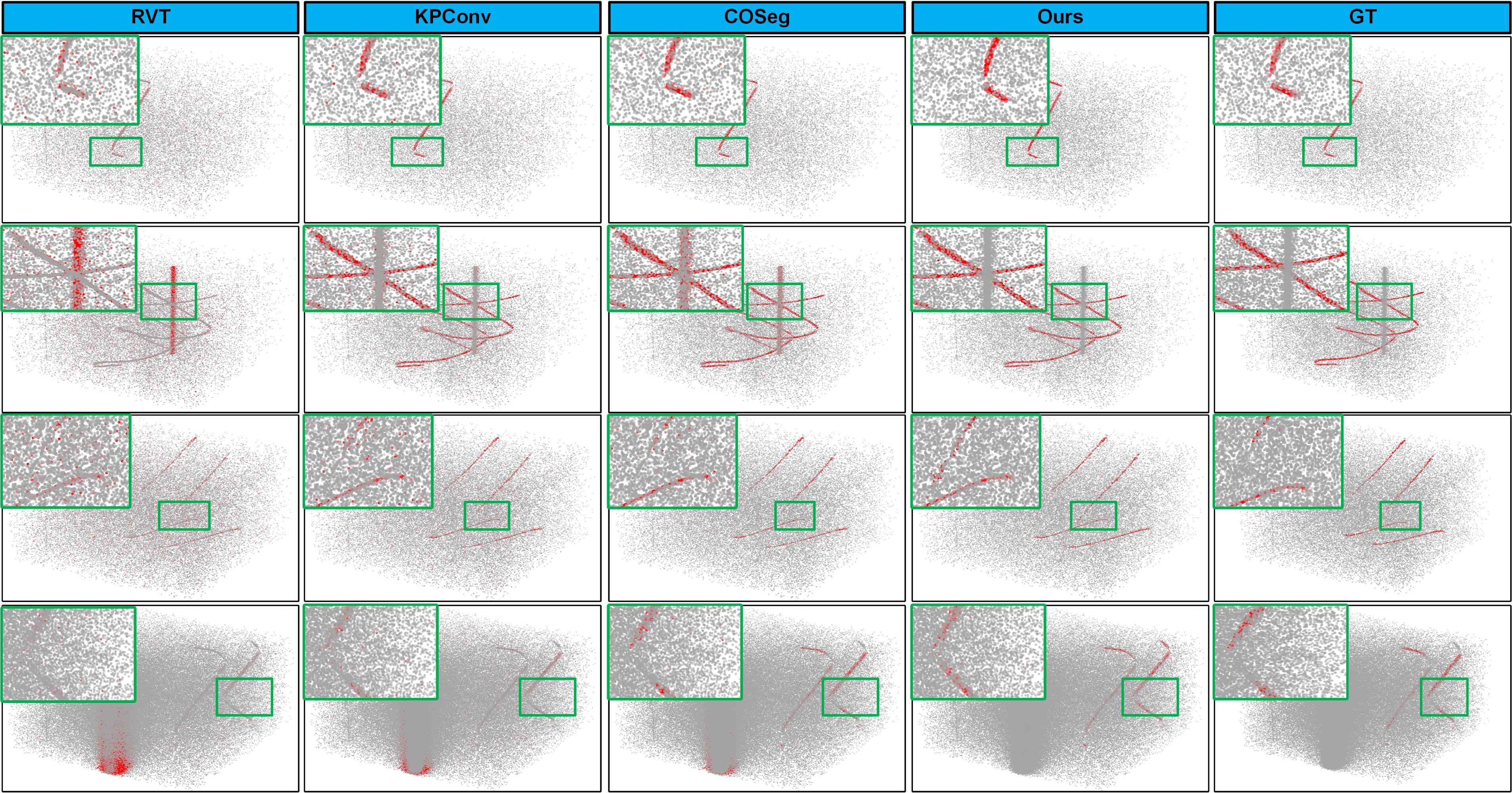}
	\caption{Qualitative results of different methods. Our method accurately detects small moving targets, while other methods produce numerous missed detections and false alarms.}
	\label{fig:vis}
\end{figure*}

\subsection{Benchmark Comparisons}
We compare the proposed method with several state-of-the-art methods, including 4 frame-based generic object detection methods (i.e., SSD~\cite{ssd}, Faster RCNN~\cite{fasterrcnn}, Deformable DETR~\cite{detr}, YOLOV10~\cite{yolov10}), 6 event based methods (i.e., RVT~\cite{rvt}, SAST~\cite{sast}, GET~\cite{get}, RED~\cite{1Mpx}, EMS-YOLO~\cite{emsyolo}, Spike-YOLO~\cite{spikeyolo}), and 3 point cloud segmentation-based methods (i.e., KPConv, 	RandLA-Net, COSeg).
We convert events within a 50 $ms$ time window into event frames~\cite{eventcamera2} as input for frame-based methods.

\textbf{Quantitative Results.} As shown in \cref{tab:sota}, our EV-SpSegNet achieve the best detection performance with  all evaluation metrics. We attribute this to two reasons. On the one hand, the GDSCA module combines local multi-scale information with global context, enabling the network to better model the differences between targets and background noise. This enhances the ability of the network  to distinguish moving targets. On the other hand, the spatiotemporal correlation loss encourages the network to identify events with high spatiotemporal correlation, improving detection performance. Additionally, by using sparse convolution to build the network, our method can inference 8 seconds of event data in 35.9 $ms$, significantly outperforming other methods.

\textbf{Qualitative Results.}  Qualitative results of different methods are shown in \cref{fig:vis}. It can be observed that our method can detect  the most moving targets in all scenes, while the compared methods exhibit more or less  missed detections and false positives. Particularly in scenes with complex backgrounds, such as the second and fourth rows with strong  noise, the improvement of our method is more significant. Much to our surprise, our method even detects some targets that were missed by manual annotation due to strong background clutter, as shown in the third and fourth rows. 
We attribute this to the spatiotemporal correlation loss, which encourages the network to retain events with high spatiotemporal correlation.

\subsection{Ablation Studies}
\subsubsection{Contribution of Components}
As shown in \cref{tab:components}, adding the GDSC block to the baseline improves the model performance in terms of $IoU$, $ACC$, and $P_{d}$ by 1.37, 2.43 and 1.18, which demonstrates that extracting multi-scale local features using GDSC is beneficial. However, adding only the PA block causes a performance drop. This is because downsampling is applied before computing global attention, and without sufficient local receptive fields, the model cannot effectively capture global features, leading to feature confusion and performance degradation.
It is worth noting that when the GDSC block and PA block are used together, the performance improves significantly. This indicates that capturing sufficient local features within patches before enabling global feature interaction is essential.  Additionally, using the STC loss  consistently improves model performance, which shows that guiding the network to retain events with high spatiotemporal correlation through STCLoss is effective.

\begin{table}[t]
	\caption{Ablation study on framework scheme and loss function.  }
	\centering
	\footnotesize
	\begin{tabular}{p{0.50cm}<{\centering}p{0.20cm}<{\centering}p{0.35cm}<{\centering}p{0.8cm}<{\centering}p{0.8cm}<{\centering}p{0.6cm}<{\centering}p{0.6cm}<{\centering}p{1cm}<{\centering}}
		\cline{1-8}
		GDSC & PA & $L_{stc}$ &  $IoU$$\uparrow$ &$ACC$$\uparrow$ & $P_{d}$$\uparrow$ & $F_{a}$$\downarrow$    &   $\#$Params. \\ \hline
		&    &         & 51.36 &60.21     &71.94       &4.81           & 5.6M     \\ \hline
		\Checkmark  &   &         & 52.73 &62.64      & 73.12      &4.28           & \textbf{3.8M}     \\
		& \Checkmark   &         & 51.26 &59.53     &71.67       & 4.31          & 5.8M       \\
		&    & \Checkmark        & 53.07 &  61.01   &  74.13     &3.66           & 5.6M     \\ \hline
		\Checkmark   & \Checkmark   &         & 53.62 &64.02     & 76.76 & 1.93     & 4.0M       \\ \hline
		\Checkmark   & \Checkmark   & \Checkmark        & \textbf{55.18} & \textbf{65.02}    & \textbf{77.53} & \textbf{1.63}      & 4.0M       \\ \hline
	\end{tabular}
		\label{tab:components}
\end{table}

\subsubsection{The Influence of the Number of Branches}
As depicted in \cref{tab:number}, we conduct an ablation experiment to study the influence of the number of branches in GDSC block on the final detection performance. 
It can be observed that the network achieves best performance when the number of branches is set to 4. This is  because too many branches result in insufficient features assigned to each channel, making it difficult to capture features at the corresponding scale. On the other hand, too few branches lead to insufficient dilation rates, preventing the network from capturing features at multiple scales.

\begin{table}[t]
	\caption{The influence of the number of  branches in GDSC block. }
	\centering
	\footnotesize
	
	\begin{tabular}{p{1cm}<{\centering}p{1cm}<{\centering}p{1cm}<{\centering}p{0.9cm}<{\centering}p{0.9cm}<{\centering}p{1cm}<{\centering}}
		\hline
		 $\#$Branches   &  $IoU$$\uparrow$ &$ACC$$\uparrow$ & $P_{d}$$\uparrow$ & $F_{a}$$\downarrow$   & $\#$Params. \\ \hline
		1     & 50.21        &   62.36                & 73.65    &2.97            &5.6M  \\
		2     & 51.41        &   64.16                & 75.21   & 2.53          &4.5M  \\
		3     & 53.52        &  64.27               & 76.66   & 2.01          &4.3M   \\
		4     & \textbf{55.18}        & \textbf{65.02}               & \textbf{77.53}    & \textbf{1.63}            & 4.0M  \\ 
		5     & 53.12       & 64.31                 & 76.53    & 2.11          &\textbf{3.8M}\\ 
		
		\hline
	\end{tabular}
	\label{tab:number}
\end{table}

\subsubsection{The Influence of Dilation Rate Combinations}
\begin{table}[t]
	\centering
	\caption{Combination of convolutions with different dilation rates. }
	\footnotesize
	\begin{tabular}{p{1.5cm}<{\centering}p{1.2cm}<{\centering}p{1.2cm}<{\centering}p{1.2cm}<{\centering}p{1.2cm}<{\centering}}
		\hline
		Dilated rates &  $IoU$($\% $)$\uparrow$ &$ACC$($\%$)$\uparrow$ & $P_{d}$($\%$)$\uparrow$ & $F_{a}$($10^{-4}$)$\downarrow$   \\ \hline
		(1,2,3,4)     & \textbf{55.18}        &\textbf{65.02}                   & \textbf{77.53}    & \textbf{1.63}               \\
		(1,2,3,5)     & 53.21        &64.11                   & 76.93    & 2.41               \\
		(1,3,5,7)     & 52.75        &64.02                   & 76.89    & 4.61            \\
		(1,3,5,9)     & 51.35        &63.92                   & 75.92    & 4.76             \\ \hline
	\end{tabular}
	\label{tab:dilation_rates}
\end{table}
We next conduct an ablation study to analyze the relationship between detection performance and different dilation rate combinations.
\cref{tab:dilation_rates} illustrates that setting the dilation rate to (1,2,3,4) achieves the best performance. 
This indicates that the dilation rate is not always better when larger. An excessively large dilation rate causes the corresponding events to be too far apart, making it difficult to effectively determine whether these events have temporal correlation.

\subsubsection{The Effectiveness of STC Loss}
To further validate the generalizability of our proposed spatiotemporal correlation Loss, we integrate it with several state-of-the-art methods, including KPConv~\cite{kpconv}, RandLA-Net~\cite{randla}, and COSeg~\cite{coseg}. As shown in \cref{tab:loss}, the addition of $L_{stc}$ consistently improves the performance of all methods. For example, KPConv with $L_{stc}$ achieves  improvements of 2.13$\%$, 1.75$\%$ and 1.4$\%$ in terms of   $IoU$, $ACC$ and $P_{d}$, respectively, and the false alarm rate decreases by 5.72$\%$.
These results show that $L_{stc}$ effectively enhances the ability of existing methods to capture spatiotemporal correlations, leading to better segmentation accuracy and reduced false alarms. The consistent performance gains across diverse methods prove the robustness and general applicability of our proposed loss function. In addition, we  explore the selection of optimal hyperparameters for the STC loss. For more details, please refer to the supplementary material.

\begin{table}[t]
	\centering
	\caption{The performance of spatiotemporal correlation loss function on different methods. }
	\footnotesize
	\begin{threeparttable}
	\begin{tabular}{p{0.9cm}|p{1.3cm}|p{1.3cm}|p{1.3cm}|p{1.5cm}}
		\hline
		Method          &  $IoU$($\% $)$\uparrow$ &$ACC$($\%$)$\uparrow$ & $P_{d}$($\%$)$\uparrow$ & $F_{a}$($10^{-4}$)$\downarrow$  \\ \hline
		KPConv    &48.19  &57.28  &68.59  &16.32                         \\
		KPConv*   &50.32(+2.13)  &59.03(+1.75)  &69.99(+1.40)  &10.60(-5.72)\\
		RandLA &50.32  &59.29  &70.56  &9.21 \\
		RandLA* &51.15(+0.83) &60.63(+1.34)  &72.36(+2.37)  &8.91(-1.69)    \\
		COSeg &51.89  &60.93  &71.32  &6.95 \\
		COSeg* &53.12(+1.23)  &62.31(0.99)  &73.96(+2.64)  &5.86(-1.09)    \\ \hline
	\end{tabular}
							\begin{tablenotes}    
		\footnotesize               
		\item  * represent the methods with STC loss.
	\end{tablenotes}  
	
	\end{threeparttable}
	\label{tab:loss}
\end{table}

\section{Discussion and Limitations}
Although our EV-UAV  accurately replicates real-world scenarios for detecting tiny UAVs in the wild, event cameras only capture motion information. When targets are stationary or move slowly, event cameras produce no output, leading to detection failures. In contrast, frame-based cameras can directly provide detailed texture information (i.e., absolute brightness) in static scenes, making them complementary to event cameras. We believe that developing a dual-modal dataset and extending our method to incorporate multi-modal inputs to maximize object detection accuracy represents a promising direction for future work.

\section{Conclusion}
In this paper, we introduce EV-UAV, the first large-scale benchmark for anti-UAV tasks. EV-UAV is a highly challenging benchmark, featuring diverse scenarios and extremely small targets. It provides event-level fine-grained annotations, supporting algorithms with different event representations. Furthermore, based on the observation that small moving targets form continuous curves in event point cloud space, we propose EV-SpSegNet, a novel baseline for segmenting target motion trajectories from sparse event point clouds, along with a Spatiotemporal Correlation (STC) loss that leverages spatiotemporal local correlation to guide the network in retaining more target events. Based on the proposed dataset, we conduct a comprehensive evaluation of our method and 13 state-of-the-art approaches, demonstrating the superiority of our method and providing a benchmark for future research in EVSOD.

{
    \small
    \bibliographystyle{ieeenat_fullname}

}

\end{document}